\documentclass[10pt,twocolumn,a4paper,fleqn]{article}
\usepackage[a4paper,left=15mm,right=15mm,top=30mm,bottom=25mm,columnsep=10mm]{geometry}
\usepackage{graphicx}
\usepackage{times}
\usepackage{titlesec}
\usepackage{indentfirst}
\usepackage[fleqn]{amsmath}
\usepackage{fancyhdr}
\usepackage{cite}
\usepackage{enumitem}
\usepackage{etoolbox}
\renewcommand{\normalsize}{\fontsize{10pt}{11.9pt}\selectfont} 
\titleformat{\section}
  {\normalfont\fontsize{11.8pt}{13.8pt}\selectfont\bfseries\MakeUppercase}{\thesection.}{0.5em}{}
\titlespacing*{\section}{0pt}{2.5ex plus 0.5ex minus .2ex}{1.5ex}
\titleformat{\subsection}
  {\normalfont\normalsize\bfseries}{\thesubsection.}{0.5em}{}
\titlespacing*{\subsection}{0pt}{1.5ex}{0.5ex}

\titleformat{\subsubsection}
  {\normalfont\normalsize}{\thesubsubsection.}{0.5em}{}
\titlespacing*{\subsubsection}{0pt}{2.0ex plus .5ex minus .2ex}{0.1ex}

\setlist[description]{
  font=\itshape,
  nosep,
  leftmargin=0pt,
  labelsep=0.5em
}

\AtBeginEnvironment{table}{
  \scriptsize
  
}

\usepackage[small,bf]{caption}
\DeclareCaptionLabelSeparator{periodspace}{.~ }
\makeatletter
\newcommand{\affils}[1]{\def\@affils{#1}}
\renewcommand{\abstract}[1]{\def\@abstract{#1}}
\newcommand{\keywords}[1]{\def\@keywords{#1}}

\renewcommand{\@maketitle}{%
  \newpage
  \null
  \begin{center}
    {\fontsize{15pt}{15pt}\selectfont \bfseries \@title \par}
    \vskip 1.0em
    {\large \@author \par}
    \vskip 0.5em
    {\normalsize \@affils \par}
  \end{center}
  {\normalsize \noindent \textbf{Abstract:} \@abstract \par}
  \vskip 1em
  {\normalsize \noindent \textbf{Keywords:} \@keywords \par}
  \vskip 1.4em
}
\makeatother

\makeatletter
\renewenvironment{thebibliography}[1]{
  \section*{\refname}
  \normalsize
  \list{[\arabic{enumi}]}{
    \settowidth\labelwidth{[#1]}
    \leftmargin\labelwidth
    \advance\leftmargin\labelsep
    \setlength{\itemsep}{0pt}
    \setlength{\parsep}{0pt}
    \setlength{\topsep}{0pt}
    \setlength{\partopsep}{0pt}
    \usecounter{enumi}
    
  }
  
  \sloppy\clubpenalty4000\widowpenalty4000
  \sfcode`\.=1000\relax
}{
  \endlist
}
\makeatother

\newcommand{\noindentbf}[1]{\noindent\textbf{#1}}
\newcommand{\mypara}[1]{\vspace{1.0mm}\noindentbf{#1}.\hspace{1em}}
\usepackage{subcaption}
\usepackage{bm}

\fancypagestyle{firstpage}{
  \fancyhf{}
  
  \fancyfoot[C]{\footnotesize \sffamily Author's version. In Proc. of the Joint Symposium of AROB 31st and ISBC 11th (AROB-ISBC 2026), pp.~1585--1590, 2026.}
}

\title{Detection and Identification of Penguins Using \\ Appearance and Motion Features }

\author{Kasumi Seko${}^{1}$, Hiroki Kinoshita${}^{2}$, Raj Rajeshwar Malinda${}^{2}$, and Hiroaki Kawashima${}^{2}$}

\affils{${}^{1}$School of Social Information Science, University of Hyogo, Hyogo, Japan\\
(E-mail: kasumi1019s@gmail.com)\\
${}^{2}$Graduate School of Information Science, University of Hyogo, Hyogo, Japan\\
(E-mail: uwuiyuzu@gmail.com, rrmalinda@gsis.u-hyogo.ac.jp, kawashima@gsis.u-hyogo.ac.jp)\\
}

\abstract{%
In animal facilities, continuous surveillance of penguins is essential yet technically challenging due to their homogeneous visual characteristics, rapid and frequent posture changes, and substantial environmental noise such as water reflections. In this study, we propose a framework that enhances both detection and identification performance by integrating appearance and motion features.
For detection, we adapted YOLO11 to process consecutive frames to overcome the lack of temporal consistency in single-frame detectors. This approach leverages motion cues to detect targets even when distinct visual features are obscured. Our evaluation shows that fine-tuning the model with two-frame inputs improves mAP@0.5 from 0.922 to 0.933, outperforming the baseline, and successfully recovers individuals that are indistinguishable in static images.
For identification, we introduce a tracklet-based contrastive learning approach applied after tracking.
Through qualitative visualization, we demonstrate that the method produces coherent feature embeddings, bringing samples from the same individual closer in the feature space, suggesting the potential for mitigating ID switching.
}


\begin{document}

\maketitle
\thispagestyle{firstpage}

\section{Introduction}

Captive animal facilities, including zoos and aquariums, serve pivotal roles in species conservation, environmental education, ecological research, and public engagement. These facilities continuously monitor animal health and behavior to understand overall animal welfare and furthermore, to improve visitor's experiences. However, long-term visual observation by keepers is labor-intensive, therefore, motivating a growing demand for automated monitoring systems~\cite{Diana2021animal}.

Penguins, as semi-aquatic species, exhibit rapid and frequent postural changes due to their terrestrial-aquatic living behavior, and therefore pose many technical challenges for automated monitoring. Among other factors, frequent occlusions between individuals occur because of their dense social nature, and extremely homogeneous visual appearance. Such similarity often causes ID switching during detection and tracking.

Accurate behavioral analysis requires uninterrupted tracking, for which high-precision detection serves as a critical foundation.
Currently, YOLO (You Only Look Once) models~\cite{hidayatullah2025review,khanam2024yolov11} demonstrate high speed and performance on still images; however, they process each frame independently and therefore cannot utilize ``temporal information'' inherent in video. Previous studies indicate that methods relying solely on appearance information perform poorly in video settings~\cite{quan2025lightweight}. In penguin monitoring, 
appearance cues are often unreliable due to several challenges, including
rapid posture changes during swimming and walking, occlusions in crowded scenes, light refraction underwater, and visual camouflage against complex backgrounds. 
Under such conditions, dynamic features such as movement and temporal change may remain more informative.

In this study, we propose a multi-frame detection approach that incorporates motion features into a still-image-based detector.
Specifically, we extend the YOLO11 model to ingest temporally adjacent frames alongside the target frame for detection. This temporal extension enables the model to learn both appearance and motion features jointly and improves detection accuracy, which we quantitatively evaluate across various configurations.
For identification, we further introduce a tracklet-based contrastive learning method applied after tracking to enhance individual discrimination, which we demonstrate through qualitative visualization.

\begin{figure}[t]
\centering
\includegraphics[width=8cm]{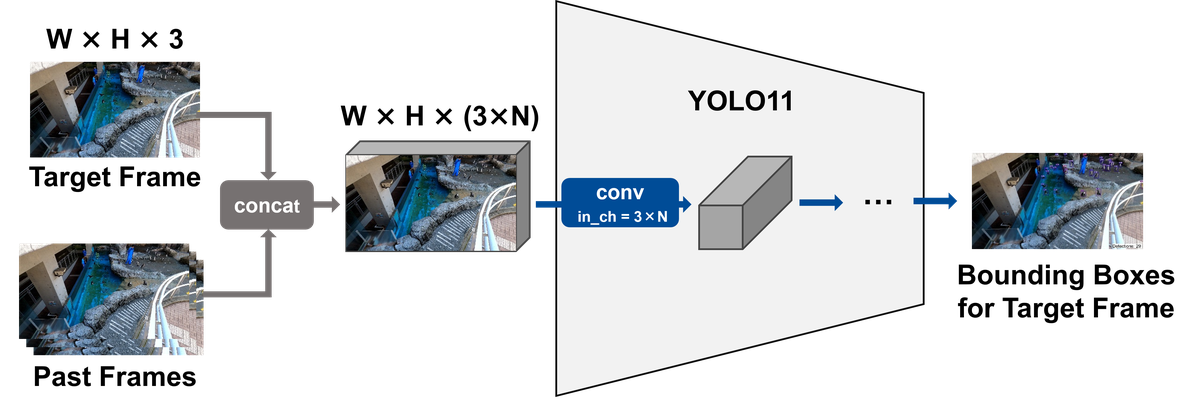}
\caption{\label{fig:Architecture} Architecture of the proposed detection method}
\end{figure}

\section{Related work}

\subsection{Detection methods utilizing temporal information}

To address the limitations caused by the lack of temporal information, various video object detection (VOD) methods have been proposed. Representative approaches include FGFA~\cite{zhu2017flow}, which propagates features from past frames based on optical flow; LSTM-based methods~\cite{lu2017online,alqaysi2021temporal} that retain time-series information; and TransVOD~\cite{zhou2023transvod}, which utilizes Transformer models to capture long-term dependencies. However, these methods are computationally expensive and may be unsuitable for applications requiring continuous, lightweight operation, such as surveillance in animal facilities.

Some alternatives integrate temporal information while maintaining lightweight characteristics. 
Quan et al.~\cite{quan2025lightweight} proposed concatenating consecutive frames in the channel dimension for YOLO, and showed improved robustness in video without significantly changing the model architecture. Similarly, Peng et al.~\cite{peng2025simple} demonstrated that incorporating inter-frame differences or optical flow into  current frame directly can effectively represent motion and lead to enhanced tracking accuracy. These methods are both computationally efficient and well-suited for lightweight VOD tasks.

\subsection{Application to animal video analysis}

In animal video analysis, incorporating temporal information is crucial for mitigating detection instability caused by crowding, occlusions, and visually similar appearances. Alqaysi et al.~\cite{alqaysi2021temporal} demonstrated that multi-frame input can significantly enhance bird detection. Similarly, Vdoviak et al.~\cite{Vdoviak2025honeybee} evaluated multiple input configurations, such as multi-frame input and inter-frame differences, for honeybee behavior analysis. Collectively, these studies emphasized the effectiveness of dynamic features.

With regard to penguins, Wu et al.~\cite{wu2023penguin} proposed a still-image-based detection method using high-resolution remote sensing images, achieving improved accuracy in wide-area population estimation. However, their method focuses exclusively on still images from a bird's-eye view, which fundamentally differs from our setting involving close-range video captured by fixed cameras in aquariums and zoos. Such close-range video includes substantial posture changes from swimming and walking, as well as frequent occlusions due to crowding. Detection methods designed solely for still images are insufficient for handling these temporal and behavioral fluctuations. Consequently, lightweight VOD for close-range, fixed-point animal observation remains an underexplored research area.

\vspace{-0.5mm}
\section{Motion-aware penguin detection}
\label{sec:detection}

Based on the approach of Quan et al.~\cite{quan2025lightweight}, we incorporate motion features by stacking multiple past frames in the channel dimension along with the target frame $I_t$, as shown in Fig.~\ref{fig:Architecture}. The output is the bounding box (bbox) for $I_t$. By expanding the input, the network automatically learns motion features (short-term appearance changes) from the initial layers. This potentially enables the detection of moving targets that are difficult to detect with still images. Additionally, we introduce inter-frame differences proposed by Peng et al.~\cite{peng2025simple} to provide motion cues to the network.

We systematically evaluated the number of input frames, frame intervals, use of inter-frame differences, and initialization methods. This evaluation clarifies the optimal configuration for penguin detection in captive facilities.

\subsection{Input configurations}

We use two configurations to select past RGB frames.

\noindentbf{Sequential Configuration (RGB-Seq)} uses temporally continuous multiple frames. The input consists of $N$ continuous images $\{  I_t, \dots, I_{t-N+1}\}$ including frame $I_t$, where $N \in \{1, \dots, 10\}$.

\noindentbf{Interval Configuration (RGB-Int)} uses the current frame $I_t$ and one past frame $I_{t-\Delta}$ spaced at an interval. The input is $\{ I_t, I_{t-\Delta} \}$, where $\Delta \in \{1, \dots, 5\}$.

To emphasize motion information, we introduce inter-frame difference images (pixel differences between two frames), referencing Peng et al.~\cite{peng2025simple}. Specifically, we define the difference image at time $t$ with interval $\Delta$ as $d_{(t, \Delta)} = \frac{I_{t+\Delta} - I_t + 255}{2}$.
Here, $I_t$ is an RGB image, and we calculate the difference for each channel. This normalizes appearance changes to a 0-255 range.
Using this difference image, we define \textbf{Difference Sequential Configuration (Diff-Seq)} and \textbf{Difference Interval Configuration (Diff-Int)} shown in Table~\ref{tab:input_types}, where both include the current frame $I_t$ along with the difference images
. %

\subsection{Initialization of model parameters}
We compare three training settings with different initialization methods for the YOLO11 pre-trained model.
\begin{description}
\item[Scratch Training:] Training from random initialization without pre-training.
\item[1st Layer Random Init:] Only the first layer is randomly initialized. Other layers retain pre-trained weights.
\item[1st Layer Replication Init:] Proposed by Quan et al.~\cite{quan2025lightweight}. The first layer filters are replicated $N$ times and scaled by $1/N$.
\end{description}

\begin{table}[t]
  \centering
  \caption{Four types of input formats defined in this study}
  \label{tab:input_types}
  \begin{tabular}{lcl}
    \hline
    Name & Abbr. & Input Content \\
    \hline
    Sequential & RGB-Seq & $\{ I_t, \dots, I_{t-N+1} \}$ \\
    Interval & RGB-Int & $\{ I_t, I_{t-\Delta} \}$ \\
    Diff-Sequential & Diff-Seq & $\{ I_t, d_{(t-1,1)}, \dots, d_{(t-N+1,1)} \}$ \\
    Diff-Interval & Diff-Int & $\{ I_t, d_{(t-\Delta,\Delta)} \}$ \\
    \hline
  \end{tabular}
\end{table}

\subsection{Experiment settings}

In order to demonstrate the effectiveness of the proposed method, we performed an evaluation experiment using the following settings.

\mypara{Dataset}
We filmed penguin flock using a fixed camera position in an aquarium, and subsequently annotated the penguins only in the video to create a dataset for this study. The frame rate of the filmed video was 29.97 frames per second (fps). This dataset contains a total of 334 training frames, 65 validation frames, and 230 test frames.

\mypara{Training Settings}
We used YOLO11m with an input size of $640 \times 640$ pixels, and optimization and data augmentation followed the default YOLO settings. Our pre-trained models, and scratch training set for 100 and 300 epochs, respectively. We selected the model with the highest mAP@0.5:0.95 in the validation data.

\mypara{Evaluation Metrics}
For evaluation metrics, we used Precision, Recall, mAP@0.5, and mAP@0.5:0.95. Furthermore, standard YOLO11 (single-frame input) was used as baseline, and further fine-tuned on our custom dataset.

\begin{table}[t]
\centering
\small
\caption{Effect of input frames $N$ in RGB-Seq ($\Delta=1$)}
\label{tab:seq_results}
\resizebox{\columnwidth}{!}{%
\begin{tabular}{c c c c c}
\hline
\textbf{Total Frames $N$} & \textbf{Precision} & \textbf{Recall} & \textbf{mAP@0.5} & \textbf{mAP@0.5:0.95} \\
\hline
\multicolumn{5}{l}{\textbf{Baseline}} \\
1 & 0.939 & 0.836 & 0.922 & 0.492 \\
\hline
\multicolumn{5}{l}{\textbf{Scratch Training}} \\
1 & 0.849 & 0.721 & 0.817 & 0.378 \\
2 & 0.866 & 0.698 & 0.790 & 0.349 \\
3 & 0.846 & 0.685 & 0.769 & 0.363 \\
5 & 0.826 & 0.709 & 0.784 & 0.361 \\
7 & 0.857 & 0.681 & 0.801 & 0.374 \\
9 & 0.862 & 0.655 & 0.775 & 0.370 \\
\hline
\multicolumn{5}{l}{\textbf{1st Layer Random Init.}} \\
1 & 0.899 & 0.768 & 0.863 & 0.442 \\
2 & 0.909 & 0.859 & 0.914 & 0.487 \\
3 & 0.958 & 0.827 & 0.920 & 0.477 \\
5 & 0.910 & 0.816 & 0.914 & 0.470 \\
7 & 0.881 & 0.808 & 0.879 & 0.438 \\
9 & 0.931 & 0.821 & 0.911 & 0.462 \\
\hline
\multicolumn{5}{l}{\textbf{1st Layer Replication Init.}} \\
1 & 0.939 & 0.836 & 0.922 & 0.492 \\
2 & 0.956 & \textbf{0.859} & \textbf{0.933} & \textbf{0.501} \\
3 & \textbf{0.961} & 0.807 & 0.910 & 0.493 \\
5 & 0.925 & 0.849 & 0.909 & 0.483 \\
7 & 0.937 & 0.845 & 0.914 & 0.488 \\
9 & 0.923 & 0.847 & 0.902 & 0.464 \\
\hline
\end{tabular}%
}
%\vspace{3mm}
\end{table}

\begin{table}[t]
\centering
\small
\caption{Effect of frame interval $\Delta$ in RGB-Int ($N=2$)}
\label{tab:int_results}
\resizebox{\columnwidth}{!}{%
\begin{tabular}{c c c c c}
\hline
\textbf{Interval $\Delta$} & \textbf{Precision} & \textbf{Recall} & \textbf{mAP@0.5} & \textbf{mAP@0.5:0.95} \\
\hline
\multicolumn{5}{l}{\textbf{Baseline}} \\
- & 0.939 & 0.836 & 0.922 & 0.492 \\
\hline
\multicolumn{5}{l}{\textbf{Scratch Training}} \\
1 & 0.866 & 0.698 & 0.790 & 0.349 \\
2 & 0.898 & 0.731 & 0.832 & 0.386 \\
3 & 0.795 & 0.725 & 0.808 & 0.355 \\
4 & 0.819 & 0.728 & 0.815 & 0.350 \\
5 & 0.850 & 0.688 & 0.785 & 0.355 \\
\hline
\multicolumn{5}{l}{\textbf{1st Layer Random Init.}} \\
1 & 0.909 & \textbf{0.859} & 0.914 & 0.487 \\
2 & 0.925 & 0.827 & 0.908 & 0.491 \\
3 & 0.913 & 0.804 & 0.886 & 0.471 \\
4 & 0.869 & 0.823 & 0.897 & 0.463 \\
5 & 0.914 & 0.819 & 0.883 & 0.456 \\
\hline
\multicolumn{5}{l}{\textbf{1st Layer Replication Init.}} \\
1 & \textbf{0.956} & \textbf{0.859} & \textbf{0.933} & \textbf{0.501} \\
2 & 0.939 & 0.845 & 0.911 & 0.481 \\
3 & 0.886 & 0.832 & 0.910 & 0.494 \\
4 & 0.920 & 0.847 & 0.909 & 0.484 \\
5 & 0.944 & 0.854 & 0.917 & 0.489 \\
\hline
\end{tabular}%
}
%\vspace{-3mm}
\end{table}

\subsection{Evaluation with RGB image input}

Tables~\ref{tab:seq_results} and~\ref{tab:int_results} show the results for RGB-Seq and RGB-Int.

\subsubsection{Effectiveness of initialization methods}
The accuracy was significantly affected by the initialization method. In all conditions tested, scratch training was not better than baseline (0.492), however, accuracy was improved by pre-trained weights. Interestingly, "1st Layer Replication Init" produced the best results. Particularly, RGB-Seq with $N=2$ and RGB-Int with $\Delta=1$ achieved the highest mAP@0.5:0.95 of 0.501.

Recall is considered a critical metric for tracking-by-detection, and our proposed method was able to achieve 0.859 in comparison to baseline (0.836). These results indicate that video information enabled the detection of individuals that are not present in still-image model.

For normal RGB frames, "1st Layer Replication Init" method was significantly effective, and likely because the pre-trained YOLO11 domain (RGB natural images) matches the input data.

\subsubsection{Influence of reference period length}
Conversely, increasing the past frames $N$ or interval $\Delta$ did not improve accuracy; rather, it caused a decline. We consider two reasons for this accuracy stagnation with longer reference periods:

First, spatial misalignment increases with the reference temporal window. Penguins move irregularly and agilely. Over a long period, the target's position changes significantly. Consequently, past frames show background or other individuals at the target's current coordinates. Since our method simply stacks these frames, irrelevant information overlaps with the target features, creating noise that hinders identification.

Second, increasing the number of input channels complicates optimization. This applies mainly to large $N$ in Sequential Configuration. Increasing $N$ directly increases input dimensions. Although immediate frame differences contain necessary motion information, excessive frames likely caused an information overload. This made it difficult for the model to extract important features.

\subsection{Evaluation with inter-frame difference input}

Tables~\ref{tab:diff_seq_results} and~\ref{tab:diff_int_results} show the results for Diff-Seq and Diff-Int.

\subsubsection{Impact of initialization on difference inputs}
Difference images produced results contrary to normal frames; "1st Layer Random Init" consistently outperformed "1st Layer Replication Init." Specifically, Diff-Int with $\Delta=2$ and Random Init achieved an mAP@0.5:0.95 of 0.501, exceeding the baseline, whereas Replication Init reached only 0.471.

We attribute this reversal phenomenon to the properties of the input data. The pre-trained model was optimized for RGB features, while inter-frame difference images differ significantly in pixel value distribution and meaning. Therefore, directly applying RGB weights (Replication Init) provided inappropriate initial values. Conversely, Random Init allowed the model to learn weights suitable for difference images from scratch, enabling the effective use of pre-trained subsequent layers.

Regarding Recall, Diff-Int with $\Delta=3$ recorded the maximum value of 0.863. This significant improvement suggests that difference information provides strong cues for preventing missed detections of moving objects.

\begin{table}[t]
\centering
\small
\caption{Effect of input frames $N$ in Diff-Seq ($\Delta=1$)}
\label{tab:diff_seq_results}
\resizebox{\columnwidth}{!}{%
\begin{tabular}{c c c c c}
\hline
\textbf{Total Frames $N$} & \textbf{Precision} & \textbf{Recall} & \textbf{mAP@0.5} & \textbf{mAP@0.5:0.95} \\
\hline
\multicolumn{5}{l}{\textbf{Baseline}} \\
1 & 0.939 & 0.836 & 0.922 & 0.492 \\
\hline
\multicolumn{5}{l}{\textbf{Scratch Training}} \\
2 & 0.914 & 0.724 & 0.842 & 0.382 \\
3 & 0.894 & 0.726 & 0.820 & 0.383 \\
5 & 0.878 & 0.688 & 0.797 & 0.369 \\
7 & 0.874 & 0.704 & 0.787 & 0.352 \\
9 & 0.889 & 0.736 & 0.826 & 0.378 \\
\hline
\multicolumn{5}{l}{\textbf{1st Layer Random Init.}} \\
2 & 0.943 & \textbf{0.842} & 0.913 & 0.481 \\
3 & 0.927 & 0.822 & 0.891 & 0.488 \\
5 & \textbf{0.950} & 0.805 & 0.894 & \textbf{0.495} \\
7 & 0.921 & 0.840 & 0.903 & 0.482 \\
9 & 0.924 & 0.837 & 0.922 & 0.464 \\
\hline
\multicolumn{5}{l}{\textbf{1st Layer Replication Init.}} \\
2 & 0.925 & 0.816 & 0.876 & 0.444 \\
3 & 0.926 & 0.839 & \textbf{0.928} & 0.473 \\
5 & \textbf{0.950} & 0.841 & 0.919 & 0.488 \\
7 & 0.895 & 0.783 & 0.869 & 0.461 \\
9 & 0.920 & 0.811 & 0.894 & 0.470 \\
\hline
\end{tabular}%
}
%\vspace{3mm}
\end{table}

\begin{table}[t]
\centering
\small
\caption{Effect of frame interval $\Delta$ in Diff-Int ($N=2$)}
\label{tab:diff_int_results}
\resizebox{\columnwidth}{!}{%
\begin{tabular}{c c c c c}
\hline
\textbf{Interval $\Delta$} & \textbf{Precision} & \textbf{Recall} & \textbf{mAP@0.5} & \textbf{mAP@0.5:0.95} \\
\hline
\multicolumn{5}{l}{\textbf{Baseline}} \\
- & 0.939 & 0.836 & \textbf{0.922} & 0.492 \\
\hline
\multicolumn{5}{l}{\textbf{Scratch Training}} \\
1 & 0.914 & 0.724 & 0.842 & 0.382 \\
2 & 0.843 & 0.725 & 0.806 & 0.366 \\
3 & 0.791 & 0.732 & 0.802 & 0.357 \\
4 & 0.878 & 0.664 & 0.790 & 0.349 \\
5 & 0.845 & 0.718 & 0.817 & 0.381 \\
\hline
\multicolumn{5}{l}{\textbf{1st Layer Random Init.}} \\
1 & \textbf{0.943} & 0.842 & 0.913 & 0.481 \\
2 & 0.935 & 0.854 & 0.921 & \textbf{0.501} \\
3 & 0.933 & \textbf{0.863} & \textbf{0.922} & 0.497 \\
4 & 0.906 & 0.795 & 0.878 & 0.462 \\
5 & 0.941 & 0.795 & 0.880 & 0.491 \\
\hline
\multicolumn{5}{l}{\textbf{1st Layer Replication Init.}} \\
1 & 0.925 & 0.816 & 0.876 & 0.444 \\
2 & 0.897 & 0.828 & 0.906 & 0.471 \\
3 & 0.919 & 0.763 & 0.852 & 0.441 \\
4 & 0.920 & 0.820 & 0.899 & 0.463 \\
5 & 0.928 & 0.763 & 0.860 & 0.444 \\
\hline
\end{tabular}%
}
\end{table}

\subsection{Qualitative evaluation}
We further performed a qualitative comparison using actual detection results. In this analysis, we used "RGB-Seq ($N=2$, Replication Init)" model, which recorded the highest accuracy and significantly improved Recall.

\subsubsection{Detection of targets with poor visual features}
\begin{figure}[t]
\centering
\includegraphics[width=8cm]{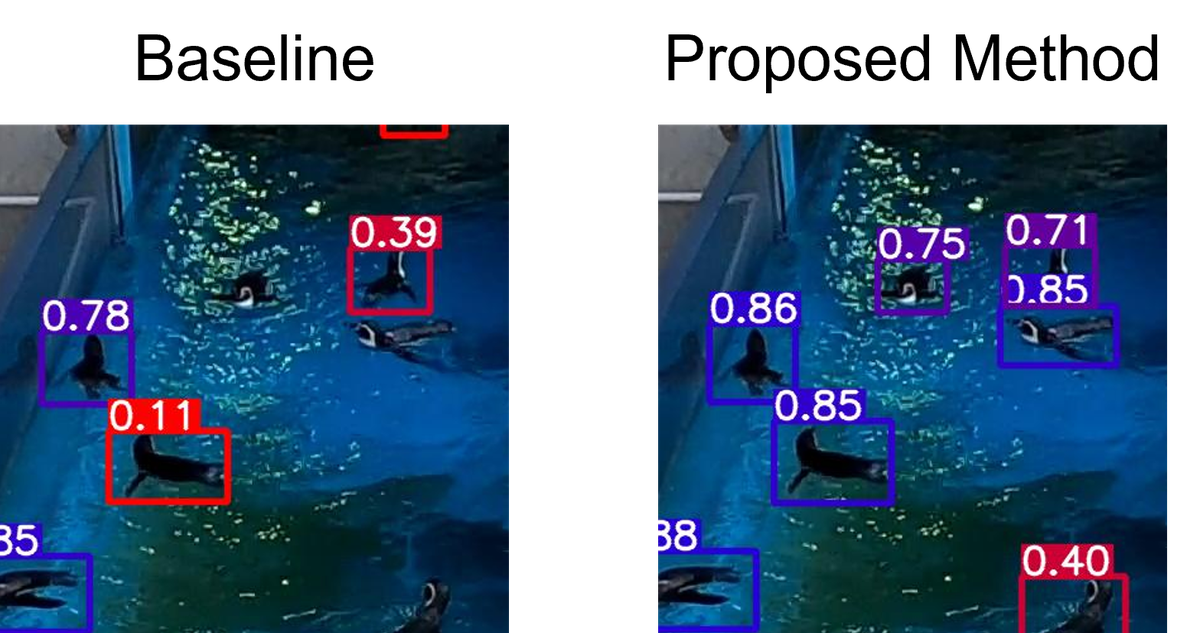}
\caption{\label{fig:qualitative_swimming} Detection results during swimming. Individuals difficult to distinguish in still images (baseline, left) are detected by utilizing video information (proposed method, right).}
\vspace{-2mm}
\end{figure}

In Fig.~\ref{fig:qualitative_swimming}, we compared the detection results for swimming penguins. In the baseline method that uses still images, light reflection and distortion on the water surface obscure the penguins. Since their bodies are mostly submerged, the baseline failed to distinguish them from the water, resulting false negatives. In contrast, the proposed method (RGB-Seq, $N=2$) effectively detected the penguins. The difference between methods, that the network likely learned water surface changes and movement information. This allowed it to use these dynamic features as cues, even when appearance features (color and shape) were unclear.

\begin{figure}[t]
\centering
\includegraphics[width=8cm]{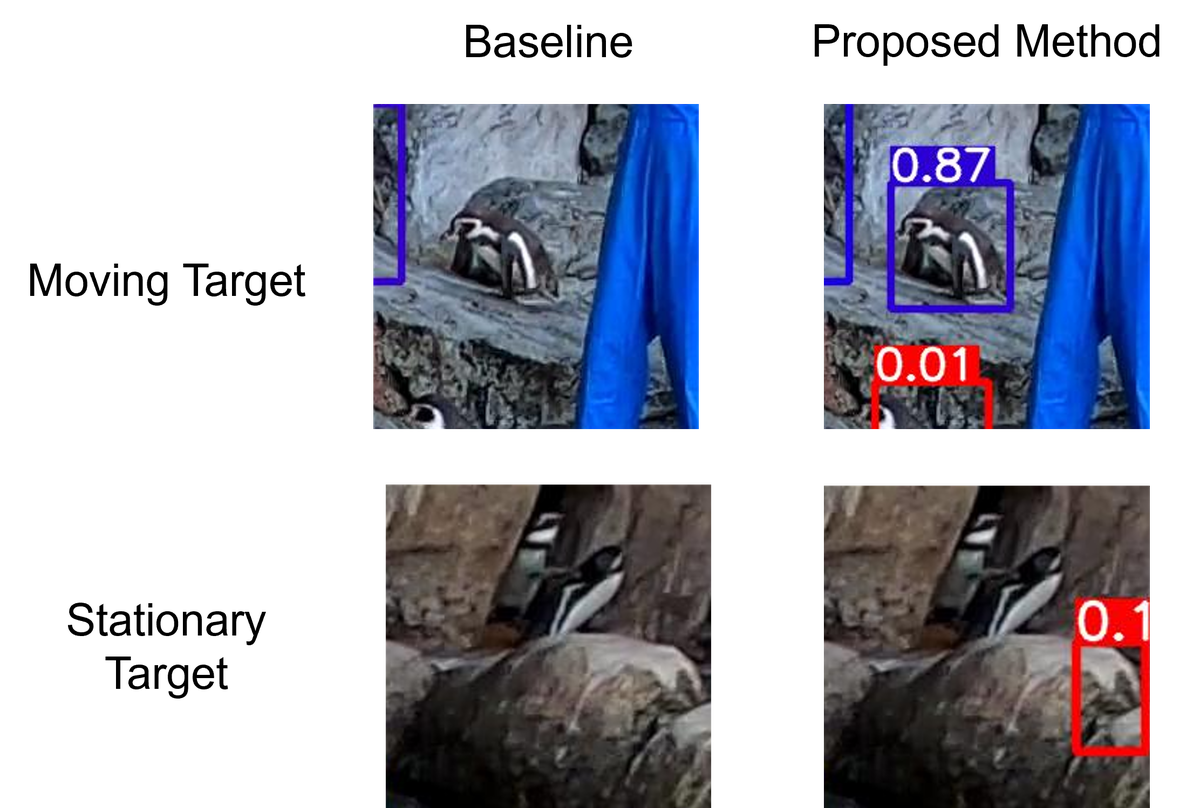}
\caption{\label{fig:qualitative_motion} Detection results in background regions unseen during training. Moving individuals were detected in both methods (top), whereas stationary individuals were not detected (bottom), illustrating the contribution of motion cues.}
\end{figure}

\subsubsection{Reduction of background dependence}
The baseline often failed in scenes where penguins appeared against backgrounds different from the training data. Models often discriminate using the background as a cue rather than the object itself. Consequently, performance decreases under new background conditions~\cite{nam2020biase}. In fixed camera video, the background remains constant, leading to excessive reliance on background context.

Figure~\ref{fig:qualitative_motion} shows an individual passing through an area with a different appearance from the training background. The baseline failed to detect the moving individual, however, the proposed method detected it correctly. Interestingly, neither of the methods detected stationary individuals in this area.

These results suggest that the baseline relied on the specific background patterns from the training. It failed because it could not infer "a background containing a penguin" in the new context. The proposed method mitigated this overfitting by utilizing motion features in addition to appearance.

\subsubsection{Limitations regarding occlusion}
Severe occlusions such as overlapping individuals were one of the limiting factor on the proposed method, and also, the accuracy was decreased in some of the cases. It is assumed that, stacking past frames most likely mixed the information from multiple individuals and background in occluded areas, that further hindered the features extraction. However, for individuals swimming alone or matching the background color, motion information proved powerful, significantly reducing missed detections. Thus, our method enhances "robustness to environmental noise (background and water)" rather than "robustness to occlusion."

\section{Learning for re-identification}
\label{sec:reid}

In the context of object tracking, a trajectory segment across consecutive frames assigned to the same identity is referred to as a \textit{tracklet} (a subsequence of the true trajectory). During tracking, tracklets are often fragmented due to occlusions or other factors, resulting in different IDs being assigned to the same individual. Therefore, re-identification (ReID) is required to associate these fragmented tracklets with their original identities.

To address this issue, this section investigates the feasibility of learning discriminative representations using appearance features of detected penguins via contrastive learning~\cite{idtracker2025}. Our objective is to train a feature encoder that minimizes the distance between features of the same individual while maximizing the distance between those of different individuals in the embedding space. Specifically, we aim to obtain an encoder capable of mapping tracklets of the same penguin to proximal feature vectors, even if they have been assigned different tracking IDs. For preliminary verification, we focus here on frame-wise feature extraction rather than modeling temporal sequences and evaluate it separately from the detection method proposed in Sec.~\ref{sec:detection}.

\subsection{Extraction of input image features}

First, we employ the pre-trained YOLO11n model and fine-tune it to adapt to the specific environment of the video footage. Subsequently, object tracking is performed to associate detected penguins across consecutive frames. A trajectory segment identified as the same individual in this process constitutes a \textit{tracklet}.

For a penguin with ID $i$, a tracklet $k$ observed during the time interval $[b_k, e_k]$ comprises the bbox sequence $(\bm{p}_{b_{k}}^{(i)},\dots,\bm{p}_{e_{k}}^{(i)})$, and we extract its corresponding appearance feature sequence $(\bm{f}_{b_{k}}^{(i)},\dots,\bm{f}_{e_{k}}^{(i)})$ using a feature extractor. 
Specifically, we utilize the Conv4 of a ResNet50 model pre-trained on ImageNet and apply RoIAlign~\cite{maskrcnn2017} to the feature maps to extract the appearance feature vector $\bm{f}_t$ corresponding to the bbox $\bm{p}_t$. %

\subsection{Self-supervised learning}

We train a multi-layer perceptron (MLP) encoder to map the extracted image features $\bm{f}_t$ into a 128-dimensional embedding space. By utilizing self-supervised learning with a contrastive loss objective, the MLP is optimized to ensure that embeddings of features belonging to the same underlying identity are proximal, while those of different identities are distant in the feature space.

Specifically, we employ triplet loss as the learning objective. The efficacy of triplet loss relies heavily on the strategy for mining positive (to be pulled closer) and negative (to be pushed away) samples relative to an anchor. In our experiments, the pool of candidate positive samples consists of all frames from tracklets sharing the same tracking ID as the tracklet containing the anchor. Conversely, candidate negative samples are drawn from all frames within tracklets that possess a different ID from the anchor and exhibit temporal overlap with the anchor's tracklet.

\subsection{Experiments and discussion}
We validate the effectiveness of the proposed method primarily through qualitative evaluation based on visualization.

\mypara{Dataset}
We utilized video footage captured by a stationary camera monitoring a penguin enclosure in an aquarium. Specifically, for this experiment, we used a clipped segment from an archived live stream provided by the Georgia Aquarium on YouTube. The subjects are African penguins (Spheniscus demersus). 
The video specifications are a resolution of $1920 \times 1080$ pixels, a frame rate of 15 fps, and a duration of 1 minute and 35 seconds.

\subsubsection{Training procedure}

We fine-tuned the YOLO model using 85 training images and 9 validation images for 100 epochs. Annotations were performed exclusively for the penguin class.
Through the tracking process across the entire video sequence, a total of 23 unique IDs and 10,997 bboxes were initially obtained. To ensure the stability of triplet loss optimization, tracklets with insufficient duration were excluded from the training set. Specifically, we treated only tracklets containing 16 or more frames as valid data. Consequently, the dataset was refined to 17 valid IDs and 10,995 bboxes.
It is important to note that identity switches caused by occlusions occurred during tracking. Through manual visual inspection, we confirmed that the following sets of IDs correspond to the same individuals: (1, 17), (15, 21), and (6, 22, 23).

For the self-supervised learning of the MLP encoder, the margin parameter for triplet loss was set to 1.0.

\subsubsection{Visualization analysis}

We visualize the embedding representations learned by the MLP using t-SNE to compare the feature distributions before and after training. Furthermore, to interpret the model's focus, we employ the method proposed by Chen et al.~\cite{gradcam2020}, which adapts Grad-CAM for deep metric learning. This allows us to visualize and identify specific regions of penguins that contribute most significantly to individual identification.

\begin{figure}[tbp]
\centering
\begin{subfigure}[t]{4cm}
\includegraphics[width=\linewidth]{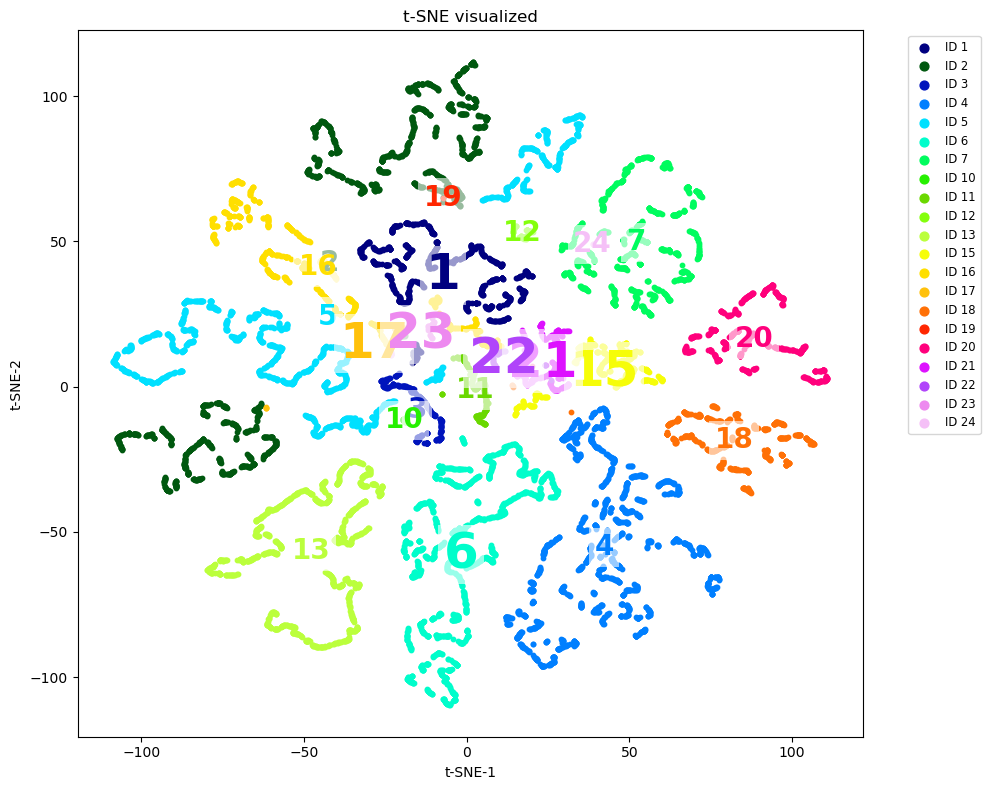}
\caption{Before}
\end{subfigure}
\begin{subfigure}[t]{4cm}
\includegraphics[width=\linewidth]{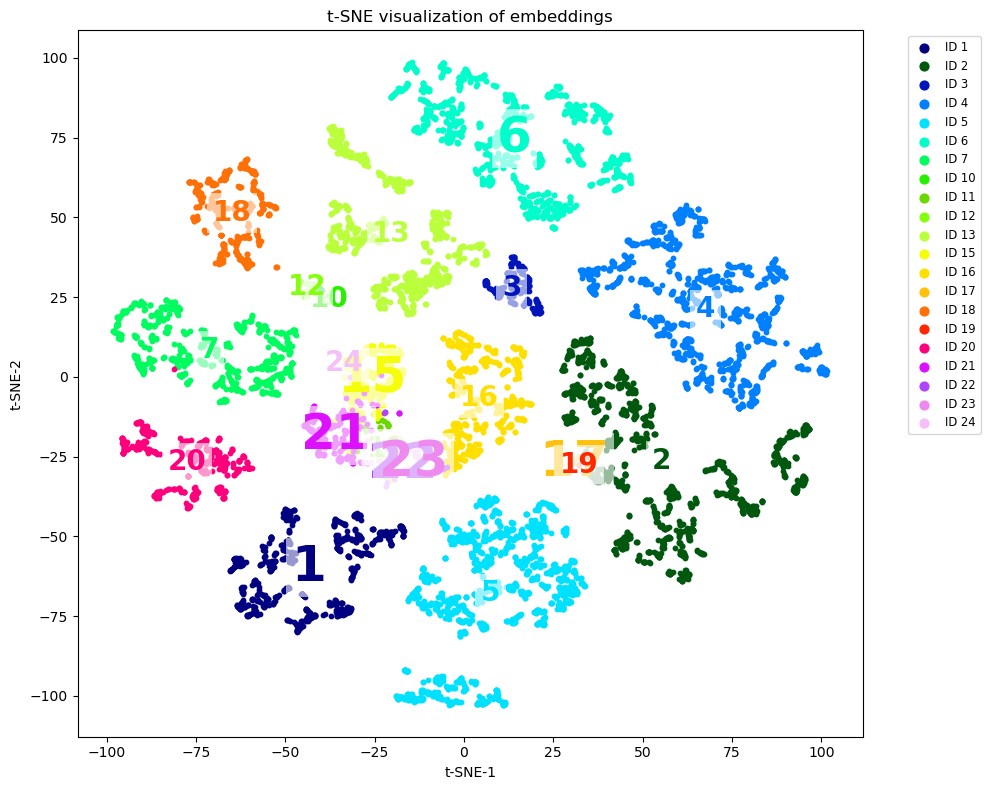}
\caption{After}
\end{subfigure}
\caption{Visualization of feature embeddings using t-SNE}
\label{fig:tsne} 
\vspace{-2mm}
\end{figure}

\mypara{t-SNE Visualization results}
Figure~\ref{fig:tsne} illustrates the visualization results before and after training. The results indicate that the training process successfully condensed the initially scattered feature points, leading to the formation of distinct clusters for each identity.
Regarding the fragmented tracklets where new IDs were assigned to the same individual during tracking (ID switches), we observed mixed results.
For the ID pair (1, 17), the clusters separated further after training.
Conversely, for the ID pair (15, 21), the clusters moved closer to each other, indicating successful learning of visual similarity. 
As for the ID group (6, 22, 23), ID 6 formed an independent cluster, whereas IDs 22 and 23 were observed to be in closer proximity.
These penguins exhibited relatively more active movement, providing diverse poses that may have helped the model learn robust identity features.

\begin{figure}[tbp]
\centering
\begin{subfigure}[t]{2cm}
\includegraphics[width=\linewidth]{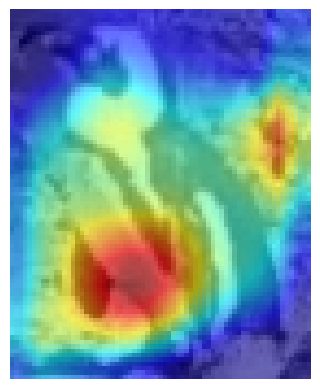}
\caption{ID15}
\label{fig:gc15}
\end{subfigure}
\hspace{5mm}
\begin{subfigure}[t]{2cm}
\includegraphics[width=\linewidth]{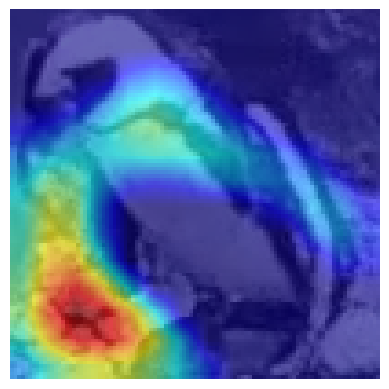}
\caption{ID21}
\label{fig:gc21}
\end{subfigure}
\caption{Grad-CAM visualization examples for IDs 15 and 21}
\end{figure}

\mypara{Grad-CAM visualization}
We present representative Grad-CAM visualization results for IDs 15 and 21 in Figs.~\ref{fig:gc15} and \ref{fig:gc21}, respectively.
In the results for ID 15, we observe that high gradients are not limited to the ventral spot patterns. While some frames show strong activation on specific body parts, others exhibit significant gradients in the background regions. A similar phenomenon is observed for ID 21; however, it exhibits an even more pronounced tendency for the model to attend to background features compared to ID 15, indicating that the model likely learned to rely on background cues rather than solely on the penguin's appearance features.

\section{Conclusion}
In conclusion, we proposed a lightweight video-based object detection method for monitoring penguins in aquarium environments through the integration of past frames into YOLO11 architecture. 
Our evaluation results demonstrate that incorporating two continuous frames with "Replication Initialization" outperformed the still-image baseline. This method effectively utilizes motion information with minimal computational resources.
Despite improved detection and accuracy, challenges such as severe occlusion in crowds are yet to be addressed. 
We also explored tracklet-based self-supervised representation learning of tracked penguins, demonstrating the potential for re-identification tasks.
Future work will focus on enhancing robustness under occlusion-heavy situations by validating the framework across diverse environmental contexts.

\section*{Acknowledgments}
We would like to thank Prof.~Yuki Kawabata at Nagasaki University and Nagasaki Penguin Aquarium (Japan) for providing the data used in Sec.~\ref{sec:detection}, and Georgia Aquarium (Atlanta, USA) for the data used in Sec.~\ref{sec:reid}.
This work was supported by JSPS KAKENHI Grant Number JP21H05302.

\end{document}